\documentclass{article} % For LaTeX2e

\usepackage{float}
\usepackage{times}
\usepackage{epsfig}
\usepackage{graphicx}
\usepackage{natbib}
\usepackage{amsmath}
\usepackage{amssymb}
\usepackage{paralist}	% compactitem
\usepackage{booktabs}
\usepackage{multirow}
\usepackage{caption}
\usepackage{subcaption}
\usepackage[percent]{overpic}

\usepackage[final]{nips_2017}
\usepackage{hyperref}
\usepackage{url}

\usepackage[usenames,svgnames]{xcolor}

\newcommand{\image}{\mathbf{d}}
\newcommand{\pose}{\mathbf{h}}
\renewcommand{\b}{\mathbf{b}}
\newcommand{\z}{{\mathbf{z}}}
\newcommand{\dkl}{D_{\textrm{KL}}}
\newcommand{\dklq}{D_{\textrm{KL}}(q(\z| \image, \pose) \| p(\z))}

\newcommand{\exppose}{\mathbb{E}_{\pose \sim q(\pose|\image)}}
\newcommand{\expz}{\mathbb{E}_{\z \sim q(\z|\image,\pose)}}

\newcommand{\expposes}{\mathbb{E}_{\pose}}
\newcommand{\expzs}{\mathbb{E}_{\z}}

\definecolor{sebastiangreen}{HTML}{006400}

\def\eg{\textit{e.g.} }
\def\ie{\textit{i.e.} }

\def\etc{\textit{etc.}}

\title{Hybrid VAE: Improving Deep Generative Models using Partial Observations}

\author{
Sergey Tulyakov\thanks{The work was done while Sergey Tulyakov was at Microsoft Research, Cambridge, UK} \\
Snap Research \\
\texttt{stulyakov@snap.com} \\
\And
Andrew Fitzgibbon, Sebastian Nowozin \\
Microsoft Research \\
\texttt{\{awf,Sebastian.Nowozin\}@microsoft.com} 
}

\begin{document}

\maketitle

\begin{abstract}
Deep neural network models trained on large labeled datasets are the
state-of-the-art in a large variety of computer vision tasks.
In many applications, however, labeled data is expensive to obtain or
requires a time consuming manual annotation process.
In contrast, unlabeled data is often abundant and available in large
quantities.
We present a principled framework to capitalize on unlabeled data by
training deep generative models on both labeled and unlabeled data.
We show that such a combination is beneficial because the unlabeled data acts
as a data-driven form of regularization, allowing generative models trained on
few labeled samples to reach the performance of fully-supervised generative
models trained on much larger datasets.
We call our method Hybrid VAE (H-VAE) as it contains both the generative and the discriminative parts.
We validate H-VAE on three large-scale datasets of different modalities:
two face datasets: (MultiPIE, CelebA) and a hand pose dataset (NYU Hand Pose). Our qualitative visualizations further support
improvements achieved by using partial observations.
\end{abstract}

\section{Introduction}
% Understanding the world -> uncertainty
Understanding the world from images or videos requires reasoning about
ambiguous and uncertain information.
For example, when an object is occluded we receive only partial information
about it, making our resulting inferences about the object class, shape, location,
or material uncertain.
% Probabilistic models
To represent this uncertainty in a coherent manner we can use probabilistic
models.
A key distinction is between \emph{generative} and \emph{discriminative}
probabilistic models, see \cite{Lasserre}.

\begin{figure}[t]
	\begin{center}
		\includegraphics[width=\textwidth]{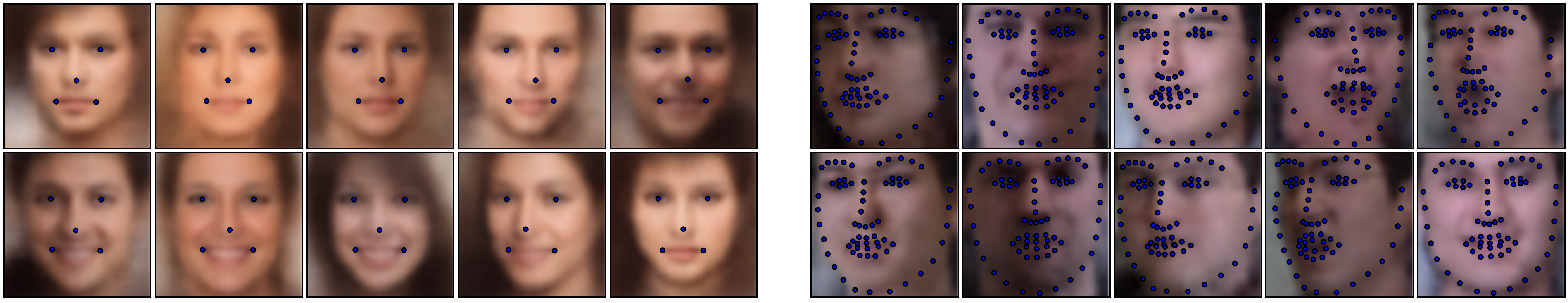}
	\end{center}
	\caption{Samples $(\image,\pose) \sim p(\image,\pose)$ of two generative models trained on the CelebA dataset (top), and on the MultiPIE dataset (bottom). Note how the models consistently sample both the image and the corresponding pose.}
	\label{fig:intro-figure}
\end{figure}

% Distinction between generative and discriminative models
\emph{Generative models} represent a joint distribution $p(\image,\pose)$ over
an observation $\image$ and a quantity $\pose$ that we would like to infer.
We can inspect a generative model by drawing samples $(\image,\pose) \sim
p(\image,\pose)$ (see Fig.~\ref{fig:intro-figure}), and we can make predictions by conditioning, evaluating
$p(\pose|\image)$.
In contrast, \emph{discriminative models} directly model the distribution
$p(\pose|\image)$, always assuming that $\image$ is observed.
We can make predictions but no longer inspect the internals of the model
through sampling.
Discriminative models often outperform generative models on prediction tasks
where a large amount of labeled data is available.
Conversely, generative models have the advantage that in principle they can
make use of abundant unlabeled data, but in practice there are computational
challenges.

% Recent progress in generative models
Today, the majority of popular computer vision models are discriminative, but recently deep learning revolutionized how we build generative models and perform inference in them.
In particular, current works on \emph{generative adversarial
networks} (GANs) (\cite{Goodfellow2014,nowozin2016fgan}) and \emph{variational autoencoders}
(VAEs)~(\cite{Kingma2013,Rezende2014,Doersch2016}) allow for rich and tractable
generative models.
In the current study, we extend the generative VAE framework to represent a joint distribution $p(\image, \pose)$ and derive a generative-discriminative hybrid making use of abundant unlabeled data.

% How? High-level key steps
To derive our hybrid model we start with a generative VAE model of the form
$p_{\theta}(\image,\pose)$, where $\theta$ are neural network parameters.
% Setting
Since labeled data is costly, we assume that only a small subset of the training 
instances is labeled $\{(\image,\pose)\}$, while a much larger training subset contains 
a collection of unlabeled observations $\{(\image)\}$ only.
% Step 1: likelihood
To allow learning from the unlabeled set we consider the \emph{marginal
likelihood} $p_{\theta}(\image) = \int p_{\theta}(\image,\pose)
\,\textrm{d}\pose$ and derive a tractable variational lower bound.
Interestingly, through a particular choice in the derivation of this lower
bound we can create an auxiliary discriminative neural network model
$q(\pose|\image)$.
% Step 2: overall objective
With the help of the bound the maximum likelihood learning objective for our
generative model now becomes the sum of the full likelihood and the marginal
likelihood.
%% Step 3: adding regression loss
%Finally, to realize trade-offs between the generative and discriminative
%models we add a third conditional likelihood term derived from the
%discriminative $q(\pose|\image)$ model, smoothly blending between a fully
%discriminative and a fully generative model.

% Benefits and contributions
The benefit of this hybrid approach is that it allows learning from partial
observations in a principled manner and scales to realistic computer vision
applications.
% Contributions
In summary, our contributions are:
\begin{itemize}
\item Deriving a principled hybrid variational autoencoder model
that allows for high-dimensional continuous output labels.
\item Using unsupervised data as data-driven regularization for large scale
deep learning models.
\item Experimentally validating the improved generative model performance in
terms of better likelihoods and improved sample quality for
facial landmarks.
\end{itemize}

\section{Full Variational Autoencoder Framework}

We extend the deep VAE framework originally presented
in~\cite{Kingma2013,Rezende2014,Doersch2016} to the case of pairs of
observations.
This extension is technically straightforward and, like the VAE approach, has
three components:
\emph{first}, a probabilistic model formulated as an infinite latent mixture
model;
\emph{second}, an efficient approximate maximum likelihood learning procedure;
and
\emph{third}, an effective variance reduction method that allows effective
maximum likelihood training using backpropagation.

% Model
For the probabilistic model, we are interested in representing a distribution
$p(\image,\pose)$.
Here $\image$ is an image, and $\pose$ is an encoding of a continuous image label,
\eg a set of locations of the markers for face alignments.
We define an infinite mixture model using an additional latent variable $\z$
as
% Infinite mixture model
\begin{equation}
p(\image, \pose) = \int p_{\theta}(\image, \pose| \z) \, p(\z) \,\textrm{d}\z.
\label{eqn:image-pose-model}
\end{equation}
The conditional distribution $p_{\theta}(\image, \pose|\z)$ is described by a
neural network and has parameters $\theta$ to be learned from a training data set.
In practice this is implemented by outputting the parameters of a multivariate
Normal distribution, $\mathcal{N}(\mu_{\theta}(\z),\Sigma_{\theta}(\z))$ so
that the conditional likelihood $p_{\theta}(\image,\pose|\z)$ can be computed
easily.
The distribution $p(\z)$ is fixed to be a multivariate standard Normal
distribution, $p(\z)=\mathcal{N}(\mathbf{0}, \mathbf{I})$.
The above model is expressive, because it corresponds to an infinite Gaussian
mixture model and hence can approximate complicated distributions. 

% Approximate MLE
To learn the parameters of the model using maximum
likelihood,~\cite{Kingma2013,Rezende2014} introduce a tractable lower-bound on
the log-likelihood.  Consider the log-likelihood of a single joint training
sample $(\image,\pose)$.  Using variational Bayesian bounding techniques
(see~\cite{Doersch2016}) we can lower bound
the log-likelihood via an auxiliary model $q(\z|\image,\pose)$ by
\begin{eqnarray}
\log p(\image, \pose)
\geq & \mathbb{E}_{\z\sim q_{\omega}(\z|\image,\pose)} \left[
	\log p_{\theta}(\image,\pose|\z)
\right]
 - \dkl(q_{\omega}(\z|\image,\pose) \| p(\z))
=: \mathcal{L}_F(\theta,\image,\pose),
\label{eqn:full-observation-likelihood}
\end{eqnarray}
where $\dkl$ is the Kullback-Leibler divergence between the variational
distribution and the prior, which for the case of two Normal distributions has
a simple analytic form.

\begin{figure}[t!]
	\centering
	\begin{overpic}[width=\textwidth]{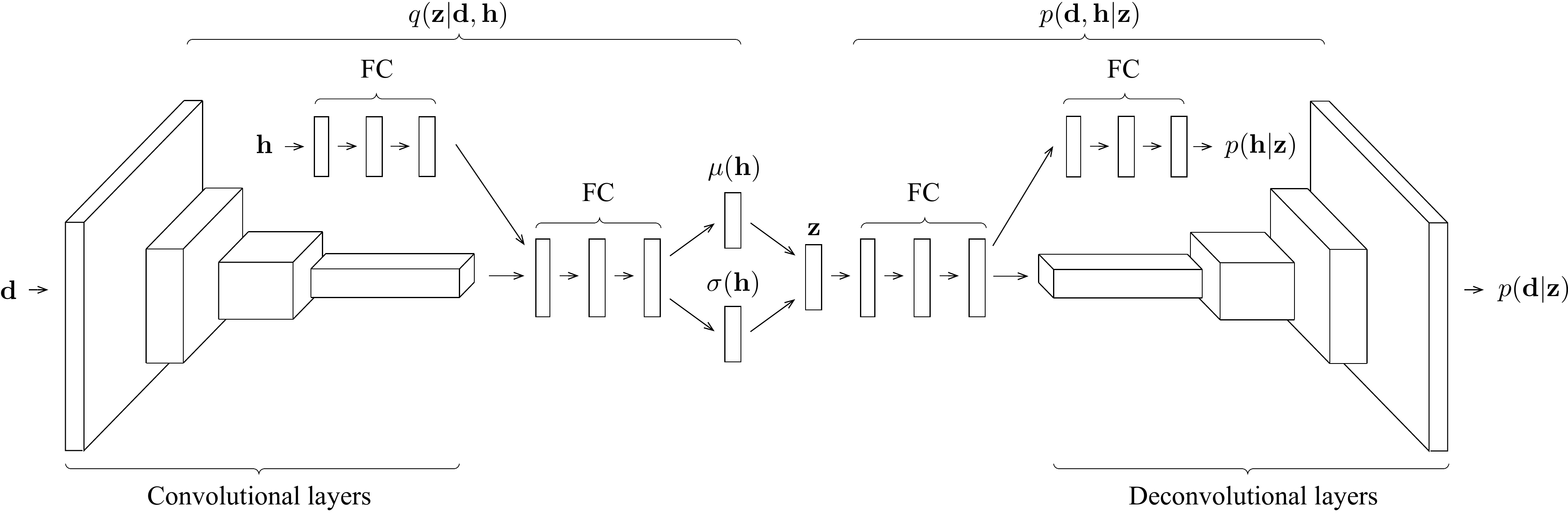}
	\end{overpic}
\caption{The architecture of the F-VAE deep generative neural
network model we use to implement
Equation~(\ref{eqn:full-observation-likelihood}).  Here FC denotes a
fully-connected neural network.}
	\label{fig:joint-vae}
\end{figure}

% Reparametrization trick
To optimize the bound~(\ref{eqn:full-observation-likelihood}) we can use stochastic
gradient descent, approximating the expectation using a few samples, perhaps
using only a single sample.
A naive sampling approach would incur a high variance in the estimated
gradients; the \emph{reparametrization trick} (see \cite{Kingma2013,Rezende2014})
allows significant variance reduction in
estimating~(\ref{eqn:full-observation-likelihood}). In the following, we refer to the models trained using eq.~\ref{eqn:full-observation-likelihood} as Full VAE or \mbox{F-VAE}, as they use only fully observed data for training. Fig.~\ref{fig:joint-vae} shows the architecture of the F-VAE model.
%%%

\section{Hybrid Learning: Using Unlabeled Data}

Obtaining a large amount of $\{\image_i\}$ is possible for many
computer vision tasks;
however, it may be expensive to collect large amounts of paired data
$(\image_i,\pose_i)$ because it involves some procedure for ground truth
collection or manual labelling of images. In the case of abundant unlabeled
data where only the image part is observed, we would like to train our model 
from \emph{both} the expensive labeled set and the partially observed data.

Given an unlabeled image $\image$, we consider the marginal likelihood
$p(\image)$ of the image as
\begin{equation}
\log p(\image) = \log \int \int p_\theta(\image, \pose|\z) p(\z)\,\textrm{d}\z \,\textrm{d}\pose.
\end{equation}
% Difficulty
The above is a difficult high-dimensional integration problem. In what follows, we drop neural network parameters $\theta$ and $w$ for brevity of notation.
Using a variational Bayes bounding technique we derive a tractable lower
bound on this marginal log-likelihood by introducing one auxiliary model,
$q(\pose|\image)$ and reusing $q(\z|\image,\pose)$ introduced in the Full VAE
framework,~(\ref{eqn:full-observation-likelihood}),
\begin{eqnarray}
\log \int \int q(\pose| \image) \frac{p(\image, \pose|\z)}{q(\pose|\image)} 
	 q(\z | \image, \pose) \frac{p(\z)}{q(\z|\image, \pose)}\,\textrm{d}\z \,\textrm{d}\pose.
\end{eqnarray}
Replacing the integrals with expectations and moving the logarithm inside
gives the variational lower bound on the log-likelihood:
\begin{eqnarray}
\log p(\image)  
\!\!\!& = &\!\!\!
\log \expposes \left[ \expzs \Big[\frac{p(\image,\pose|\z)}{q(\pose|\image)} 
	\frac{p(\z)}{q(\z|\image, \pose)} \Big]  \right]	
\geq 
\expposes \expzs \left[\log \frac{p(\image,\pose|\z)}{q(\pose|\image)} + 
	\log \frac{p(\z)}{q(\z|\image, \pose)} \right] \nonumber\\
\!\!\!& = &\!\!\!
\expposes \Big[\expzs \Big[ \log p(\image,\pose|\z) - \log q(\pose|\image) \Big] - \dklq \Big].
\label{eqn:pdbound-1}
\end{eqnarray}
Here, $\expposes$ is shorthand for $\exppose$, and likewise $\expzs$ stands for
$\expz$.
We rewrite~(\ref{eqn:pdbound-1}) by recognizing the entropy term $H(q(\pose|
\image))=-\expzs[\log q(\image| \pose)]$, giving
\begin{eqnarray}
\log p(\image)
\geq 
\expposes\Big[
	\expzs \left[\log p_{\theta}(\image,\pose|\z)\right] - \dklq \Big] + H(q(\pose| \image)) \, 
 =: \mathcal{L}_P(\theta,\image).
\label{eqn:pdbound-2}
\end{eqnarray}
In our case, the entropy $H(q(\pose|\image))$ is available as a simple
analytic form because we use a multivariate Normal distribution for
$q(\pose|\image)$. Note that $q(\pose| \image)$ represents essentially a discriminative model,
implemented using a deep neural network.

We now combine $\mathcal{L}_F$ and $\mathcal{L}_P$ into one learning
objective.
For this, we assume we have a dataset $\{(\image_i,\pose_i)\}_{i=1,\dots,n}$
of fully-observed samples and another dataset $\{(\image_j)\}_{j=1,\dots,m}$
of partially-observed data, so that only $\image$ is observed.
Typically $m \gg n$ because it is easier to obtain unlabeled images.
Because both $\mathcal{L}_F$ and $\mathcal{L}_P$ are log-likelihood bounds for
a single instance, one principled way to combine the two learning objectives
is to simply sum them over all instances,
\begin{equation}
\mathcal{L}_1(\theta) :=
	\sum_{i=1}^n \mathcal{L}_F(\theta,\image_i,\pose_i)
	+ \sum_{j=1}^m \mathcal{L}_P(\theta,\image_j).
	\label{eqn:L1}
\end{equation}
While~(\ref{eqn:L1}) is a valid log-likelihood bound, we found that
empirically learning is faster when the relative contribution of each
sum is weighted equally.
We achieve this through the learning objective
\begin{equation}
\mathcal{L}(\theta) :=
	\frac{1}{n} \sum_{i=1}^n \mathcal{L}_F(\theta,\image_i,\pose_i)
	+ \frac{1}{m} \sum_{j=1}^m \mathcal{L}_P(\theta,\image_j).
	\label{eqn:L}
\end{equation}
We optimize~(\ref{eqn:L}) using minibatch stochastic gradient descent,
sampling one separate minibatch for each sum per iteration. We consider $\mathcal{L}(\theta)$ in (\ref{eqn:L}) to be a hybrid learning objective, as it couples together generative and discriminative models. Hence the name of the model: Hybrid VAE (H-VAE).

\begin{figure}[t!]\centering%
	\hfill%
	\begin{subfigure}[b]{0.2\linewidth}%
		\includegraphics[width=0.99\linewidth]{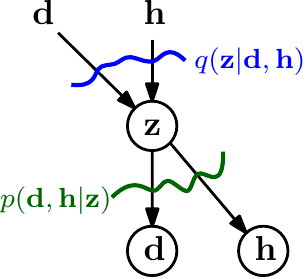}%
		\caption{Full VAE}
		\label{fig:vae-standard}
	\end{subfigure}%
	\hfill%
	\begin{subfigure}[b]{0.2\linewidth}%
		\includegraphics[width=0.99\linewidth]{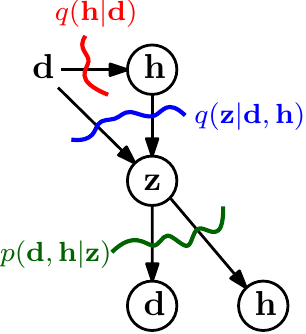}%
		\caption{Hybrid VAE}
		\label{fig:vae-semi}
	\end{subfigure}%
	\hfill%
	\begin{subfigure}[b]{0.4\linewidth}
	\includegraphics[width=\linewidth]{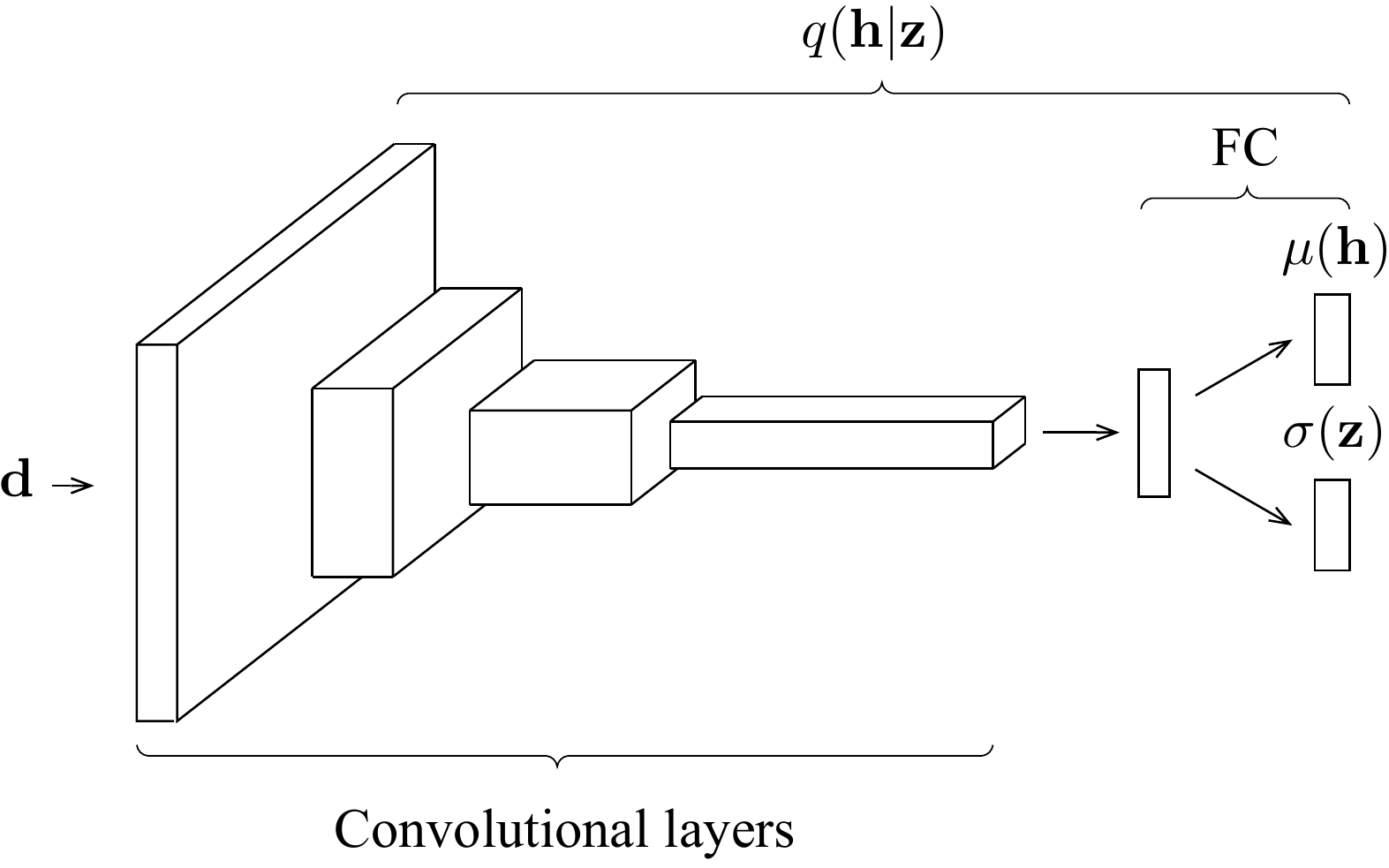}		
        \caption{Architecture of $\color{red} q(\pose| \image)$}
	\label{fig:discriminative-network}
	\end{subfigure}
	\hfill%
	\vspace{-0.2cm}%
	\caption{Extending the variational autoencoder (VAE) model to the case of
		hybrid labeled/unlabeled observations.
		% (a)
		(\subref{fig:vae-standard}) the standard VAE model extended to the paired
		observation case: an encoder {\color{blue}$q(\z|\image,\pose)$} and decoder
		{\color{DarkGreen}$p(\image,\pose|\z)$} mapping to/from a latent code $\z$;
		% (b)
		(\subref{fig:vae-semi}) the hybrid VAE model introducing a discriminative
		variational model {\color{red}$q(\pose|\image)$}
		% (c)
		(\subref{fig:discriminative-network}) The architecture of the discriminative network.
	}
	\label{fig:vae-network}
\end{figure}

We show in the experimental section that hybrid learning using~(\ref{eqn:L})
greatly improves the log-likelihood on the hold-out set of the generative
model.  Additionally we show that the use of a large set of partially observed
instances prevents overfitting of the generative models. Fig.~\ref{fig:vae-standard} and Fig.~\ref{fig:vae-semi} outline the components of the F-VAE and H-VAE frameworks respectively.

\subsection{Modeling Depth Images}
Compared to natural images, depth images have the additional property that
depth values could be \emph{unobserved}.  This is either because a pixel is
outside the operating range of the camera or because the pixel is invalidated
by the depth engine.

To model this effect accurately, we proceed in two steps: for each pixel, we
first compute a probability of being observed, $p(\b|\z)$, where $\b$ is a
probability map with one probability $b_u \in [0,1]$ for each position $u$.  If
$b_u=1$ then the normal continuous model $p(d_u|\z)$ is used, but if $b_u=0$,
then the depth value is set to $d_u=\emptyset$, a symbolic ``unobserved'' value.
Formally this corresponds to enlarging the domain of depth observations to
$\mathbb{R} \cup \{\emptyset\}$ and using the probability model
$p(\image|\z) = \int p(\image|\b,\z) \, p(\b|\z) \,\textrm{d}\b$.
We can implement this simple model efficiently as a summation over two maps
because there are only two states for each pixel.

\section{Experiments}

\subsection{Implementation Details}

An H-VAE model has three neural networks as shown in Fig.~\ref{fig:vae-semi}.
Each of these networks uses either convolutional or deconvolutional
subnetworks, and these parts closely follow the encoder-decoder architecture
proposed in~\cite{radford2015unsupervised}.
We visualize the network architectures in Fig.~\ref{fig:joint-vae} and
Fig.~\ref{fig:discriminative-network}.
Every convolutional layer doubles the number of channels, while shrinking the
width and height by a factor of two.
Deconvolutatal layers perform the opposite operation: they reduce the number
of channels by the factor of two, but double the width and height by a factor
of two. 

Given a pair of $(\image, \pose)$ the encoding network $q(\z|\image, \pose)$
independently processes $\image$ using the convolutional subnetwork, while
$\pose$ is processed by the three-layer fully-connected neural network
(FC-NN). These two outputs are then concatenated and passed through another
FC-NN, producing a diagonal multivariate Normal distribution over $\z$.
The decoder network $p(\image, \pose|\z)$ first processes $\z$ using an FC-NN,
the pipeline then split, and the deconvolutional subnetwork is used to produce
a diagonal multivariate Normal distribution over $\image$.
The distribution over $\pose$ is computed by yet another FC-NN, again as a
diagonal multivariate Normal.
We use ReLU activations~(\cite{nair2010relu}) throughout all the networks and
every FC-NN hidden layer has 256 units.
In every stochastic layer, we use unbiased estimates for the expectations by
averaging three samples from the corresponding distribution.
We implement the pipeline in \emph{Chainer} (\cite{tokui2015chainer}) and train
it end-to-end using SGD with learning rate $0.01$ and momentum $0.9$.

\subsection{Datasets}
Previous work on generative models used data sets such as MNIST~(\cite{lecun1998mnist})
and SVHN~(\cite{netzer2011reading}) for evaluating their models.
In our work, however, these datasets do not allow us to show the benefits of the
proposed H-VAE approach, as they include \emph{categorical} labels only,
whereas our method extends the standard VAE framework to deal with
\emph{continuous} annotation $\pose$.

We evaluate our method on two up-to-date datasets used in the computer vision
community. To simulate partially observed samples, we shuffle the dataset and
split it into fully- and partially-observed subsets and separate a holdout
test set, not available during training.

\noindent{\textbf{MultiPIE.}} The MultiPIE (\cite{gross2010multi}) dataset consists of face images of 337 subjects taken under different pose, illumination and expressions. The pose range contains 15 discrete views, capturing a face profile-to-profile. Illumination changes were modeled using 19 flashlights located in different places of the room. The database has been extensively used in the community for face alignment (\cite{xiong2013supervised, zhu2012face, tulyakov2017pami}). For our purposes we use only the views annotated either with 68 or 66-points markup, in total producing 47250 images. We drop the inner mouth corner points, so that all the images have the same 66-point markup. Images were cropped around the landmarks and downscaled to 48$\times$48 size. We reserve 2K $(\image, \pose)$ pairs from the MultiPIE dataset for testing purposes.

\noindent{\textbf{CelebA face dataset.}} The CelebA (\cite{liu2015deep}) is a
large scale face attributes dataset containing more that 200K celebrity images
annotated with 40 attributes (such as eyeglasses, pointy nose \etc) and 5
landmark locations (eyes, nose, mouth corners). It contains more than 10K
distinct identities.
As in~\cite{lamb2016discriminative} we center and crop all images around the face and resize to $64\times64$. We use the provided landmarks as continuous labels $\pose$. The testing set consists of 10K fully observed instances.

\noindent{\textbf{NYU Hand Pose dataset.}} The NYU Hand Pose Dataset (\cite{tompson14tog}) contains 72757 training frames of RGBD data and 8252 testing set frames. For every frame, the RGBD data from 3 kinects is available. In our experiments we use only depth frames captured from the front view. We resize depth maps to $128\times128$ and preprocess the depth values using the code from~\cite{oberweger2015hands}.

\subsection{Quantitative Evaluation}

\begin{table}
	\centering
	\caption{Comparison of F-VAE and H-VAE models on three datasets: (\subref{tab:exp-multipie}) MultiPIE, (\subref{tab:exp-celeba}) CelebA, (\subref{tab:exp-nyu}) NYU Hand Pose. The number of fully observed samples used for training is denoted by $n$, while $m$ corresponds to the total number of partially observed instances. We report the task-loss $E_\text{task}$ and the negative log-likelihood of the fully observed testing data computed using eq.~\ref{eqn:full-observation-likelihood}.}
	\resizebox{\textwidth}{!}{%
% 	\resizebox{\textwidth}{!}{%
		\begin{minipage}{0.5\textwidth}
			\begin{subtable}{\linewidth}
				\centering
				\subcaption{MultiPIE}
				\label{tab:exp-multipie}
				\begin{tabular}{cllcc}
					\toprule
					{} &  $n$ &  $m$  &  $E_\text{task}$ & $-\log p(\image, \pose)$ \\
					\midrule
					\parbox[t]{2mm}{\multirow{3}{*}{\rotatebox[origin=c]{90}{F-VAE}}}
					& 5k   & -  & - & -23.00 \\
					& 15k  & -  & - & -27.33 \\
					& 30k  & -  & - & -31.25 \\
					\midrule
					\parbox[t]{2mm}{\multirow{4}{*}{\rotatebox[origin=c]{90}{H-VAE}}}
					&  500 & 5k   & 0.1162 & -20.43 \\
					&  500 & 30k  & 0.1457 & -25.75 \\
					&  5k  & 5k   & 0.0935 & -28.29 \\
					&  5k  & 30k  & 0.0845 & \textbf{-32.36} \\
					\bottomrule
				\end{tabular}
			\end{subtable}
		\end{minipage}
% 		}
		%
% 		\resizebox{\textwidth}{!}{%
		\begin{minipage}{0.5\textwidth}
			\begin{subtable}{\linewidth}
				\centering
				\subcaption{CelebA}
				\label{tab:exp-celeba}	
				\begin{tabular}{cllcc}
					\toprule
					{} &  $n$ &  $m$  &  $E_\text{task} $ & $-\log p(\image, \pose)$ \\
					\midrule
					\parbox[t]{2mm}{\multirow{5}{*}{\rotatebox[origin=c]{90}{F-VAE}}}
					& 15k  & - & - & -3.49 \\
					& 30k  & - & - & -7.40 \\
					& 50k  & - & - & -7.19 \\
					& 100k & - & - & -8.15 \\
					\midrule
					\parbox[t]{2mm}{\multirow{5}{*}{\rotatebox[origin=c]{90}{H-VAE}}}
					& 5k  & 100k  & 0.1425 & -6.90  \\
					& 5k  & 150k  & 0.1420 & -6.98  \\
					& 15k & 15k   & 0.1181 & -5.90  \\
					& 15k & 100k  & 0.0929 & \textbf{-9.08} \\
					\bottomrule
				\end{tabular}
			\end{subtable}
		\end{minipage}
% 		}
		%
% 		\resizebox{\textwidth}{!}{%
		\begin{minipage}{0.5\textwidth}
			\begin{subtable}{\linewidth}
				\centering
				\subcaption{NYU Hand Pose}
				\label{tab:exp-nyu}	
				\begin{tabular}{cllcc}
					\toprule
					{} &  $n$ &  $m$  &  $E_\text{task} $ &  $-\log p(\image, \pose)$ \\
					\midrule
					\parbox[t]{2mm}{\multirow{2}{*}{\rotatebox[origin=c]{90}{F-VAE\hspace{-5.5pt}}}}
					\rule{0pt}{3ex}   
					& 10k  & - & -  & 29.31 \\
					& 30k  & - & -  & 26.27 \\  
					\midrule
					\parbox[t]{2mm}{\multirow{6}{*}{\rotatebox[origin=c]{90}{H-VAE}}}
					& 10k  & 10k   & 4.54  & 29.31  \\
					& 10k  & 20k   & 4.82  & 25.89  \\
					& 10k  & 30k   & 4.56  & 21.13  \\
					& 30k  & 10k   & 4.82  & 19.20  \\
					& 30k  & 20k   & 4.56  & 18.42  \\
					& 30k  & 30k   & 4.56  & {\bf 14.11}  \\
					\bottomrule
				\end{tabular}
			\end{subtable}
		\end{minipage}%
% 		}
	}
	\label{table:experiments}
\end{table}

A commonly accepted comparison procedure for generative modeling is evaluation
of the negative log-likelihood (NLL) on the testing
data~(\cite{gneiting2007scoringrules,lamb2016discriminative,Kingma2016}). We compute the NLL using
Eq.~(\ref{eqn:full-observation-likelihood}) on the same fully observed testing
data for each model. This metric, however, is difficult to interpret,
requiring visual samples from the model to assess their quality. Therefore, we
additionally report the \emph{task}-loss $E_\text{task}$. The task-loss in the case of face
alignment is the average point-to-point Euclidean distance, normalized by the
interocular distance~(\cite{tulyakov2015regressing, trigeorgis2016mnemonic}). For hand pose experiment we report the average $L_2$ distance between the ground truth and the prediction. We
report the task-loss by evaluating the mean of $q(\pose| \image)$ trained using the hybrid
objective~(\ref{eqn:L}). This metric is not available for the F-VAE models. For brevity reasons, we denote \mbox{F-VAE($n$)} as a
F-VAE model trained using $n$ fully observed samples, and similarly
\mbox{H-VAE($n$, $m$)} trained with $n$ fully- and $m$ partially-observed
instances.

Table~\ref{tab:exp-multipie} compares multiple F-VAE models against the proposed H-VAE models on the MultiPIE dataset. Clearly, using a hybrid learning objective with partially observed data helps drastically improve the likelihood: the H-VAE(500, 30K) model outperforms the F-VAE(5K). Similarly, the H-VAE(5K, 30K) model shows better NLL as compared to F-VAE(30K). 

The same holds for the CelebA dataset, as seen in table~\ref{tab:exp-celeba}.
The F-VAE(5K) model is not able to accurately learn the distribution.  Instead, it
overfits, leading to the worse NLL. In contrast, the \mbox{H-VAE(5K, 15K)} has
comparable results with the F-VAE(15K), advocating for the use of inexpensive
partial observations. Additionally, H-VAE(30K, 150K) outperforms F-VAE(150K). Similar results can be observed on the NYU Hand Pose dataset (Table \ref{tab:exp-nyu}), where the H-VAE(10k, 10k) models shows NLL comparable to F-VAE(30K), and clearly the model having the most of fully observed and partially observed data scores best.

\subsection{Qualitative Evaluation}

\begin{figure}[t!]
	\begin{subfigure}[b]{\textwidth}
		\centering
		\includegraphics[width=0.95\textwidth]{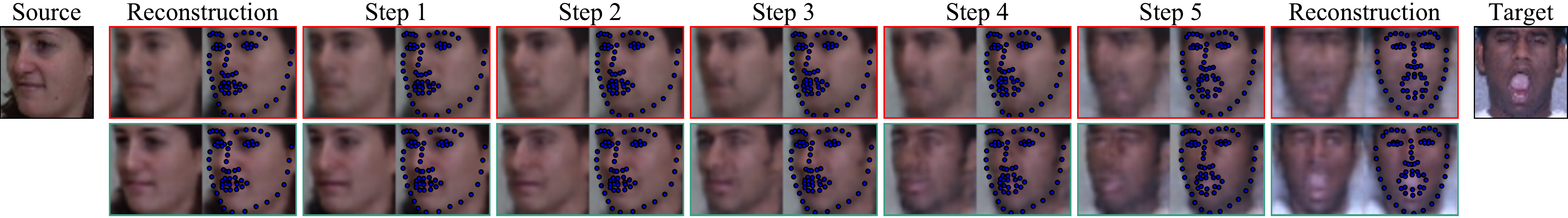}
		\caption{Samples the first row (red) for every example are obtained using the F-VAE(5K) model, the bottom row (green) is produced by the H-VAE(5K, 30K) model. Both models were trained on the MultiPIE dataset.}
			\label{fig:mpie-interpolation}
	\end{subfigure}
	\begin{subfigure}[b]{\textwidth}
	\centering
	\includegraphics[width=0.95\textwidth]{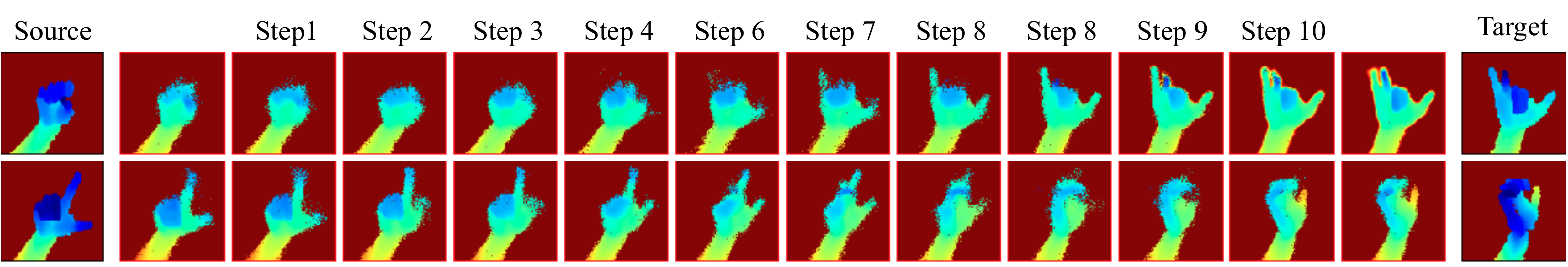}
	\caption{Interpolating depth images using the H-VAE(30K, 30K) model trained on the NYU Hand Pose dataset.}
		\label{fig:nyu-interpolation}
	\end{subfigure}
	\caption{Selected examples showing interpolations. For every interpolation, the source and the target images are taken from the testing set, projected to and reconstructed from the latent space, interpolation is performed from the source to the target. }
\end{figure}

One of the key benefits of generative modeling is the ability to analyze the
learned distribution by performing sampling. Since $\z \sim
\mathcal{N}(\mathbf{0}, \mathbf{I})$ one can sample $\z$ and then sample
$(\image, \pose) \sim p(\image, \pose|\z)$ to get an image and its label.
Fig.~\ref{fig:intro-figure} shows several examples of that. Note how a
generative model can consistently represent pose and image information. The
CelebA dataset consists of public images of celebrities, and therefore is
biased towards smiling faces. The bottom half of Fig.~\ref{fig:intro-figure}
shows samples from the distribution learned using the MultiPIE dataset. Since
the MultiPIE dataset consists of multiple poses, illuminations and
expressions, the model is able to encode them. Interestingly, one can project
two images onto the $\z$-space by sampling $\z \sim q(\z|\image, \pose)$ and
analyze the structure of the learned space by linearly interpolating between the two points. 

Source-target interpolation on the MultiPIE dataset is given in Fig.~\ref{fig:mpie-interpolation}. The
models were trained on the MultiPIE dataset. For this example, the top
row is obtained using the F-VAE(5K) model that saw only 5K labeled examples during
training. The bottom row is produced using the model trained
on 5K fully observed samples and 30K of partial observations (H-VAE(5K, 30K)).
Clearly, partial observations significantly improve image quality, making the
texture less blurry, and better representing facial expressions.
Interestingly, both models gradually transform between images (\ie open a mouth or rotate a
face), indicating that by exploring the latent representation one can generate
plausible face trajectories.  It also shows that, in general, useful semantics
are \emph{linearized} in the hidden representation of the model.

Similar interpolation on the NYU Hand Pose dataset is given in Fig.~\ref{fig:nyu-interpolation}. Interestingly, gesture transformation is encoded in the representation. Traversing a line transforms to bending or unbending fingers in the image space.

\section{Related work}
\label{section:related-work}

We first discuss existing state-of-the-art deep generative models.
Since our formulation is joint in terms of images and their labels we review
previous attempts to create such a joint representation.
Additionally, as our model is a hybrid consisting of discriminative and
generative parts we outline previous related works in this area.

\subsection{Deep Generative Models}
\label{section:deep-generative-models}
Current research focuses on two classes of models,
\emph{generative adversarial networks} and \emph{variational autoencoders}.

\textbf{Generative adversarial networks (GAN)} were proposed recently
in the machine learning community as neural architectures for generative
models~(\cite{Goodfellow2014}).
In a GAN two networks are trained together, competing against each other:
the \emph{generator} tries to produce realistic samples, \eg images;
the \emph{adversary} tries to distinguish generated samples from training
samples.
Formally, this yields a challenging $\min-\max$ optimization problem and a
variety of techniques have been proposed to stabilize learning in
GANs, see \cite{salimans2016improved,sonderby2016gansuperresolution,metz2016unrolledgan}.

Despite this difficulty, multiple extensions of the original work appeared in
the literature. Deep convolutional generative adversarial network
in~\cite{radford2015unsupervised} and~\cite{denton2015lapgan} show
surprisingly photo-realistic and sharp image samples as compared to previous
works.
The authors provide multiple architectural guidelines that improve the overall
quality of the sampled images. As a result the models are able to produce
samples for trajectories in the latent space, showing the internal structure
of the learned space.
The work in~\cite{nowozin2016fgan} shows that GANs can be viewed as a special
case of a more general variational divergence estimation approach.
Further examples of GANs include
generating images of birds from textual description, \cite{reed2016generative},
styling images, \cite{ulyanov2016texture}, and
video generation, \cite{vondrick2016generating, tulyakov2017mocogan}.

\textbf{Variational autoencoders} were introduced independently by two
groups~(\cite{Kingma2013,Rezende2014}).
VAEs maximize a variational lower bound on the log-likelihood of the data.
Similarly to GANs, this is a recently emerged and rapidly evolving area of
generative modeling. 

Since the original works, there have been many extensions introduced.
The work in~\cite{Kingma2014} extends the VAE framework to semi-supervised
learning.
In~\cite{Maaloe2015} the idea of semi-supervised learning is exploited
further, by introducing and auxiliary variables improving the variational
approximation.
The Deep Recurrent Attentive Writer (DRAW), \cite{gregor2015draw}, employs
recurrent neural networks acting as encoder and decoder, in a way mimicking
the foveation of the human eye.
Inverse Autoregressive Flow (IAF) presented in \cite{Kingma2016} and auxiliary
variables~(\cite{Maaloe2015}) are two approaches to further improve the quality
of the variational approximation and quality of the images generated by the
model.
Another recent attempt at improving inferences is to consider hybrid GAN-VAE
models as in~\cite{wu2016objectshapes}.

\subsection{Hybrid Models}
\label{section:hybrid-models}
Learning a probabilistic model from different levels of annotations has been
proposed earlier in~\cite{navaratnam2007jointmanifoldmodel}; in particular,
using Gaussian processes (GP) the authors report improved person tracking
performance.
However, while GPs are analytically tractable they do not scale well and the
work is limited to use very small data sets.

In~\cite{Lasserre} the authors consider the problem of combining a generative
models $p(\image,\pose)$ with a discriminative model $p_d(\pose|\image)$.
They achieve this in a satisfying manner by creating a new model whose
likelihood function can smoothly balance between training objectives of the
generative and discriminative models.
However, the proposed coupling prior of~\cite{Lasserre} is not useful in the
context of neural networks because Euclidean distance in the parameter vector
of two neural networks does not measure useful differences in the function
realized by the neural network.
The model is shown to work well for discrete labels $\pose$ in the empirical
study~(\cite{druck2007semisupervised}).

Whereas the above work addresses semi-supervised learning with the goal to
improve the predictive performance of $p(\pose|\image)$, our main interest is
in improving the performance of the generative model $p(\image,\pose)$.
Moreover, our work shows to to handle the marginal likelihood over $\pose$ in
an efficient ant tractable manner.

\section{Conclusions}

We demonstrated a scalable and practical way to learn rich generative models
of multiple output modalities.
Compared to the ordinary deep generative model our hybrid VAE model does not
require fully labelled observations for all samples.
Because the hybrid VAE model derives from the principled variational
autoencoder model it can represent complex distributions of images and pose
and could be easily adapted to other modalities.

Our experiments demonstrate that when mixing fully labelled with unlabelled
data the hybrid learning greatly improves over the standard generative model
which can use only the fully labelled data.
The improvement is both in terms of test set log-likelihood and in the
quality of image samples generated by the model.

We believe that hybrid generative models such as our hybrid VAE model address
one of the \emph{key limitation of deep learning}: the requirement of having
large scale labelled data sets.
We hope that hybrid VAE model will enable large scale learning from
unsupervised data for a variety of computer vision tasks.

{\small
\bibliographystyle{plainnat}
\bibliography{paper}
}

\end{document}